\documentclass[9pt]{article}

\usepackage{spconf,amsmath,graphicx,}
\usepackage{algorithm2e}
\usepackage{amsmath}
\usepackage{amssymb}
\usepackage{booktabs}
\usepackage{tabularx}
\usepackage{hyperref}
\hypersetup{
    colorlinks=true,
    linkcolor=black,
    citecolor=black,
    urlcolor=blue,
}
\usepackage{multirow}

\usepackage{cite}
\title{Unsupervised Contrastive Hashing for Cross-Modal Retrieval in Remote Sensing}
\name{Georgii~Mikriukov, Mahdyar~Ravanbakhsh, Begüm~Demir}
\address{Technische Universität Berlin, Berlin, Germany}
\begin{document}
\ninept
\maketitle
\begin{abstract}

 The development of cross-modal retrieval systems that can search and retrieve semantically relevant data across different modalities based on a query in any modality has attracted great attention in remote sensing (RS). In this paper, we focus our attention on cross-modal text-image retrieval, where queries from one modality (e.g., text) can be matched to archive entries from another (e.g., image). Most of the existing cross-modal text-image retrieval systems in RS require a high number of labeled training samples and also do not allow fast and memory-efficient retrieval. These issues limit the applicability of the existing cross-modal retrieval systems for large-scale applications in RS. To address this problem, in this paper we introduce a novel unsupervised cross-modal contrastive hashing (DUCH) method for text-image retrieval in RS. To this end, the proposed DUCH is made up of two main modules: 1) feature extraction module, which extracts deep representations of two modalities; 2) hashing module that learns to generate cross-modal binary hash codes from the extracted representations. We introduce a novel multi-objective loss function including: i) contrastive objectives that enable similarity preservation in intra- and inter-modal similarities; ii) an adversarial objective that is enforced across two modalities for cross-modal representation consistency; and iii) binarization objectives for generating hash codes. Experimental results show that the proposed DUCH outperforms state-of-the-art methods. Our code is publicly available at \url{https://git.tu-berlin.de/rsim/duch}.

\end{abstract}
\begin{keywords}
cross-modal retrieval, hashing, unsupervised contrastive learning, remote sensing.
\end{keywords}
%

\section{Introduction}
\label{sec:intro}

The increased number of recent Earth observation satellite missions has led to significant growth in remote sensing (RS) image archives. Thus, the development of scalable and accurate retrieval systems (which aim at searching for semantically similar data to a given query) for massive archives is one of the most important research topics in RS. Most existing methods in RS focus on content-based RS image retrieval (CBIR). CBIR systems take a query image and compute the similarity function between the query image and all archive images to find the most similar images to the query \cite{sumbul2021deep}. To achieve scalable image retrieval, deep hashing techniques have become a cutting-edge research topic for large-scale RS image retrieval \cite{roy2021hashing, Cheng2021hashing}. These methods map high-dimensional image descriptors into a low-dimensional Hamming space where binary hash codes describe the image descriptors. Compared with the real-valued features, hash codes allow fast image retrieval by calculating the Hamming distances with simple bit-wise XOR operations. In addition, the binary codes can significantly reduce the amount of memory required for storing the content of images. 

The abovementioned methods are defined for single-modality image retrieval problems (called uni-modal retrieval). For a given query image, uni-modal retrieval systems search for the images with semantically similar contents from the same modality image archive \cite{cao2020enhancing} (e.g., Sentinel-2 images). However, multi-modal data archives, including different modalities of satellite images as well as textual data, are currently available. Thus, the development of retrieval systems that return a set of semantically relevant results of different modalities given a query in any modality (e.g., using a text sentence to search for RS images) has recently attracted significant attention in RS. Due to the semantic gap among different modalities, it is challenging to search relevant multi-modal contents among heterogeneous data archives. To address this problem, cross-modal retrieval methods that aim to identify relevant data across different modalities are recently introduced in RS \cite{chen2020deep, cheng2021deep, ning2021semantics}. The existing cross-modal retrieval systems in RS are defined based on supervised retrieval techniques. Such techniques require the availability of labeled training samples (i.e., ground reference samples) to be used in the learning phase of the cross-modal retrieval algorithm. The amount and the quality of the available training samples are crucial for achieving accurate cross-modal retrieval. Collecting a sufficient number of reliable labeled samples is time-consuming, complex and costly in operational scenarios, significantly affecting the final accuracy of cross-modal retrieval. 

\begin{figure*}[t]
  \centering
  \includegraphics[width=\linewidth]{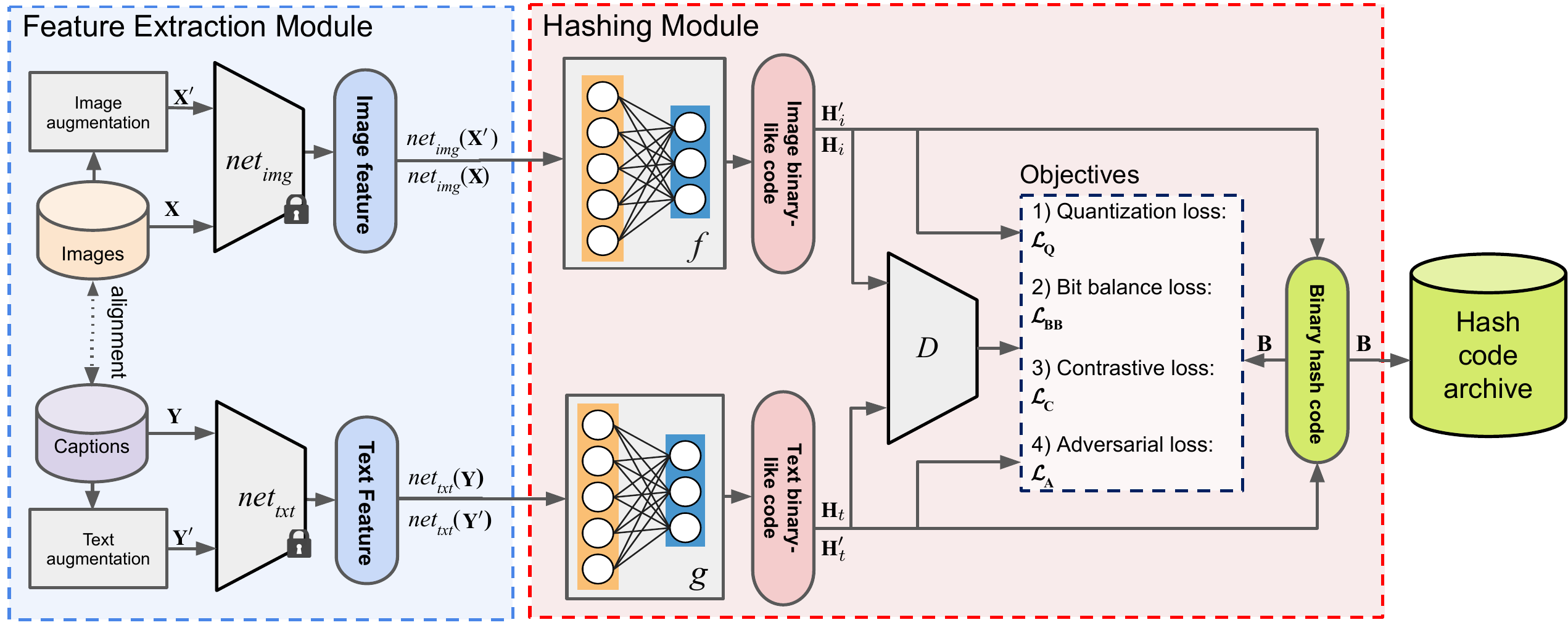}
  \caption{
Block diagram of the proposed method. In the feature extraction module, the modality-specific encoders $net_{img}$ and $net_{txt}$ extract deep embedding for image and text modalities, respectively. The hashing module learns two hash functions $f$ and $g$ from the input embedding. Discriminator $D$ enforces the adversarial objective between two modalities for cross-modal representation consistency.}
  \label{fig:main-diagram}
\end{figure*}

Unlike RS, in the computer vision (CV) community, unsupervised and in particular self-supervised cross-modal representation learning methods (which only rely on the alignments between modalities) are widely studied \cite{jung2021contrastive,su2019deep,liu2020joint, chen2020improved,chen2020simple,he2020momentum}. As an example, in \cite{su2019deep} a deep joint-semantics reconstructing hashing (DJSRH) method is introduced to learn binary codes that preserve the neighborhood structure in the original data. To this end, DJSRH learns a mapping from different modalities into a joint-semantics affinity matrix. Liu et al. \cite{liu2020joint} propose a joint-modal distribution-based similarity weighting (JDSH) method based on DJSRH, exploiting an additional objective based on cross-modal semantic similarities among samples. The existing unsupervised contrastive learning methods mainly rely on cross-modal (inter-modality) contrastive objectives to obtain consistent representations across different modalities, while the intra-modal contrastive objectives are ignored. This may lead to learning an inefficient embedding space, where the same semantic content is mapped into different points in the embedding space \cite{zolfaghari2021crossclr}.

In this paper, we focus on the cross-modal retrieval between RS image and text sentence (i.e., caption), which consists of two sub-tasks: i) RS image retrieval with a sentence as query; and ii) sentence retrieval with an RS image as the query. In detail, we propose a novel deep unsupervised cross-modal contrastive hashing (DUCH) method that considers both inter- and intra-modality contrastive objectives for representation learning. In detail, we introduce a novel multi-objective loss function that consists of: 1) intra- and inter-modal contrastive objectives that enable similarity preservation within and between modalities; 2) an adversarial objective enforcing cross-modal representation consistency; 3) binarization objectives to generate representative binary hash codes.

\section{Proposed Method}
\label{sec:method}




We assume that an unlabelled multimodal training set $\textbf{O}$ that consists of $N$ number of image and text pairs (i.e. samples) is available, where $\textbf{O} = \{\textbf{X}, \textbf{Y}\}^N$ and each image is described with only one caption. $\textbf{X} = \{\textrm{x}_m\}_{m=1}^N$ and $\textbf{Y} = \{\textrm{y}_m\}_{m=1}^N$ are image and text modality sets, respectively, where $\textrm{x}_m \in \mathbb{R}^{d_i}$ and $\textrm{y}_m \in \mathbb{R}^{d_t}$ are image and text feature vectors. Image and text feature vector dimensions are defined by $d_i$ and $d_t$. Given the multimodal training set $\textbf{O}$, the proposed DUCH aims at learning hash functions $f$ and $g$ for image and text modalities, respectively. In detail, the hash functions $f$ and $g$ learn to generate binary hash codes $\textbf{B}_i = f(\textbf{X},\theta_i)$ and $\textbf{B}_t = g(\textbf{Y},\theta_t)$, where $\textbf{B}_i \in \{0 , 1\}^{N \times B}$ and $\textbf{B}_t \in \{0 , 1\}^{N \times B}$ for image and text modalities. $\theta_i$, $\theta_t$ are parameters of image and text hashing networks and $B$ is the length of binary hash code. 
In order to learn the hash functions $f(.)$ and $g(.)$, the proposed DUCH includes two main modules: 
i) the feature extraction module that produces deep semantic representations for image and text modalities and input them to the next module; and ii) the hashing module that generates binary representations from the input extracted deep features. Fig. \ref{fig:main-diagram} shows the block diagram of the DUCH, indicating the two modules. In the following, we describe each module in detail.


\subsection{Feature extraction module}
\label{sec:method-components-feature}


The feature extraction module aims at generating deep semantic representations for both images and captions, and feed them into the next module (hashing). To this end, the feature extraction module includes two networks as modality-specific encoders: 1) an image embedding network denoted as $net_{img}$; and 2) a language model for encoding the textual information (i.e., captions) denoted as $net_{txt}$. Weights of image and text encoders are fixed during the training of the hashing learning module. Given the training set $\textbf{O}$, the image and text embedding are extracted by $net_{img}(\textbf{X})$ and $net_{txt}(\textbf{Y})$, respectively. For the sake of simplicity in the rest of this paper we refer $net_{img}(\textbf{X})$ as $\textbf{X}$, and $net_{txt}(\textbf{Y})$ as $\textbf{Y}$. For the unsupervised contrastive representation learning of DUCH, we generate a corresponding augmented set from $\textbf{O}$, which is defined as $\textbf{O}^\prime = \{\textbf{X}^\prime, \textbf{Y}^\prime\}^N$, where $\textbf{X}^\prime = \{\textrm{x}_m^\prime\}_{m=1}^N$ and $\textbf{Y}^\prime = \{\textrm{y}_m^\prime\}_{m=1}^N$ are augmented image and caption with elements $\textrm{x}_m^\prime \in \mathbb{R}^{d_i}$ and $\textrm{y}_m^\prime \in \mathbb{R}^{d_t}$. The embedding of augmented images and captions are extracted by $net_{img}$ and $net_{txt}$, respectively. For the sake of simplicity in the rest of this paper we refer $net_{img}(\textbf{X}^\prime)$ as $\textbf{X}^\prime$, and $net_{txt}(\textbf{Y}^\prime)$ as $\textbf{Y}^\prime$.

\subsection{Hashing module}
\label{sec:method-components-hash}


The hashing module aims at learning hashing function to generate cross-modal binary hash codes from the image and text embeddings. To this end, we employ multiple objectives for training hash functions $f$ and $g$, including contrastive loss, adversarial loss and binarization losses. The cross-modal contrastive loss is the main objective for unsupervised representation learning in the proposed DUCH. In detail, we introduce an intra-modal (modality-specific) contrastive loss to improve the learned representations. The adversarial objective ensures the modality-invariance of the binary codes. The binarization objectives are required to approximate better the generated continuous values close to the discrete binary codes.

\noindent\textbf{Contrastive objectives}. In order to calculate the contrastive losses we use the normalized temperature scaled cross entropy objective function adapted from \cite{chen2020simple, zhang2021cross}. To ensure better representation learning we consider both inter-modal and intra-modal contrastive losses. The former maps features of both modalities into a common space, while the latter performs the mapping within each modality. The inter-modal contrastive loss $\mathcal{L}_{C_{inter}}$ is defined as:
\begin{align}
\label{eq:contrastive-loss-inter}
\small
    &\mathcal{L}_{C_{inter}}(\textrm{x}_j,\textrm{y}_j) =\\
    &-\log \frac{S\left( f(\textrm{x}_j), g(\textrm{y}_j) \right) }{ \sum_{k=1, k\neq j}^{M} S\left( f(\textrm{x}_j), f(\textrm{x}_k) \right) + \sum_{k=1}^{M} S\left( f(\textrm{x}_j), g(\textrm{y}_k) \right) }, \nonumber
\end{align}
where $S(\textrm{u},\textrm{v}) = \exp\left( \cos\left( \textrm{u},\textrm{v} \right) / \tau \right)$, and $\cos\left( \textrm{u}, \textrm{v} \right) = \textrm{u}^T\textrm{v}/\left\| \textrm{u} \right\|\left\| \textrm{v} \right\|$ is the cosine similarity, $\tau$ denotes a temperature, and $M$ is a batch size. The inter-modal contrastive loss $\mathcal{L}_{C_{inter}}$ is computed between image $\textrm{x}_j$ and its paired caption $\textrm{y}_j$. The intra-modal contrastive losses for image and text modalities are computed as: 
\begin{align}
\label{eq:contrastive-loss-intra-img}
\small
    &\mathcal{L}_{C_{img}}(\textrm{x}_j,\textrm{x}^\prime_j) =\\
    &-\log \frac{S\left( f(\textrm{x}_j), f(\textrm{x}^\prime_j) \right) }{ \sum_{k=1, k\neq j}^{M} S\left( f(\textrm{x}_j), f(\textrm{x}_k) \right) + \sum_{k=1}^{M} S\left( f(\textrm{x}_j),f(\textrm{x}^\prime_k) \right) }, \nonumber\\ \nonumber\\
    &\mathcal{L}_{C_{txt}}(\textrm{y}_j,\textrm{y}^\prime_j) =\\
    &-\log \frac{S\left( g(\textrm{y}_j), g(\textrm{y}^\prime_j) \right) }{ \sum_{k=1, k\neq j}^{M} S\left( g(\textrm{y}_j), g(\textrm{y}_k) \right) + \sum_{k=1}^{M} S\left( g(\textrm{y}_j),g(\textrm{y}^\prime_k) \right) }, \nonumber
\end{align}
where $\mathcal{L}_{C_{img}}$ is the contrastive loss between image $\textrm{x}_j$ and its augmentation $\textrm{x}^\prime_j$, while $\mathcal{L}_{C_{txt}}$ is the contrastive loss between caption $\textrm{y}_j$ and its augmentation $\textrm{y}^\prime_j$. The final contrastive loss $\mathcal{L}_{C}$ is defined as: 
\begin{equation}
\small
    \mathcal{L}_{C} = \mathcal{L}_{C_{inter}} + \lambda_1 \mathcal{L}_{C_{img}} + \lambda_2 \mathcal{L}_{C_{txt}},
    \label{eq:contrastive-loss}
\end{equation}
where $\lambda_1$ and $\lambda_2$ are hyperparameters for image and text intra-modal contrastive losses, respectively.

\noindent\textbf{Adversarial objective}. As suggested in \cite{bai2020deep}, to enforce the consistency of representations across modalities, we employ an adversarial loss within DUCH. We use a discriminator network $D$ trained in an adversarial fashion, where the text embedding is assumed to be ``real'' and the image embedding is considered as ``fake''. The adversarial objective function $\mathcal{L}_{adv}$ is defined as:
\begin{align}
\label{eq:adversarial-loss}
\small
    \mathcal{L}_{adv}\left( \mathbb{X}, \mathbb{Y} \right) = -\frac{1}{N}\sum^{N}\Big[ &\log\Big( D\big( g \left( \mathbb{Y} \right) \big) \Big) +\\
    &\log \Big( 1 - D\big( f(\mathbb{X}) \big) \Big) \Big],\nonumber
\end{align}
where $\mathbb{X} = \{\textbf{X},\textbf{X}^\prime\}$ for the image representations and $\mathbb{Y} = \{\textbf{Y},\textbf{Y}^\prime\}$ for the text representations. $D(\textbf{*}, \theta_D)$ $(* =\mathbb{X}, \mathbb{Y})$ is the discriminator network $D$ with parameters $\theta_D$. 

\noindent\textbf{Binarization objectives}. In order to generate representative hash codes, we consider two main objective functions for binarization: quantization loss \cite{jiang2017deep} and bit balance loss \cite{yang2017pairwise}. The former optimizes the difference between continuous and discrete hash values, and the latter enforcing each output neuron to fire with an equal chance. 
The quantization loss $\mathcal{L}_{Q}$, and bit balance loss $\mathcal{L}_{BB}$ are calculated as: 
\begin{equation}
\label{eq:binarization-quant}
\small
    \mathcal{L}_{Q} = \left\| \textbf{B} - \textbf{H}_i \right\|^2_F +
    \left\| \textbf{B} - \textbf{H}_i^\prime \right\|^2_F +
    \left\| \textbf{B} - \textbf{H}_t \right\|^2_F +
    \left\| \textbf{B} - \textbf{H}_t^\prime \right\|^2_F,
\end{equation}
\begin{equation}
\small
    \mathcal{L}_{BB} = \left\| \textbf{H}_i \cdot \textbf{1} \right\|^2_F +
    \left\| \textbf{H}_i^\prime \cdot \textbf{1} \right\|^2_F +
    \left\| \textbf{H}_t \cdot \textbf{1} \right\|^2_F +
    \left\| \textbf{H}_t^\prime \cdot \textbf{1} \right\|^2_F,
    \label{eq:binarization-bit-balance}
\end{equation}
where $\textbf{H}_i = f( \textbf{X})$, $\textbf{H}_i^\prime = f( \textbf{X}^\prime)$, $\textbf{H}_t = g( \textbf{Y})$, $\textbf{H}_t^\prime = g( \textbf{Y}^\prime)$ are binary like codes for images, augmented images, texts and augmented texts respectively, and $\textbf{1}$ is N-dimensional vector with all values of 1. The final binary code update rule is defined as:
\begin{equation}
\small
    \textbf{B} = sign \left( \frac{1}{2} \left( \frac{\textbf{H}_i + \textbf{H}_i^\prime}{2} + \frac{\textbf{H}_t + \textbf{H}_t^\prime}{2} \right) \right).
    \label{eq:binarization-final-binary-code}
\end{equation}
The overall loss function is a weighted sum of multiple objectives from \eqref{eq:contrastive-loss}, \eqref{eq:adversarial-loss}, \eqref{eq:binarization-quant}, and \eqref{eq:binarization-bit-balance}:
\begin{equation}
\small
    \min_{\textbf{B}, \theta_i, \theta_t, \theta_D} \mathcal{L} = \mathcal{L}_C + \alpha \mathcal{L}_{adv} + \beta \mathcal{L}_Q + \gamma \mathcal{L}_{BB},
    \label{eq:loss-overall}
\end{equation}
where $\alpha$, $\beta$, $\gamma$ are hyperparameters for adversarial, quantization and bit balance losses, respectively. Finally, to retrieve semantically similar captions to a query image $\textrm{x}_q$, we compute the Hamming distance between $f(net_{img}(\textrm{x}_q))$ and hash codes in the retrieval archive. The obtained distances are ranked in ascending order and the top-$K$ captions with the lowest distances are retrieved. Similarly, for a query caption $\textrm{y}_q$, the Hamming distance between $g(net_{txt}(\textrm{y}_q))$ and hash codes in the retrieval archive are computed, ranked and top-$K$ images are retrieved.


\section{Experimental Results}
\label{sec:experiments}


In the experiments we used the RSICD dataset \cite{lu2017exploring} that includes $10921$ images of $31$ classes from aerial orthoimagery. Each image has a size of $224 \times 224$ pixels and has $5$ corresponding captions. Only one randomly selected caption for each image was used during the training. The dataset was split by random selection into the train, query, and retrieval sets (50\%, 10\% and 40\%, respectively). The image augmentation was performed by applying a Gaussian blur filter with kernel size $3 \times 3$ and $\sigma \in [1.1, 1.3]$, random rotation in the range of $[-10^\circ, -5^\circ]$, and $200 \times 200$ center cropping. For the text augmentation, we selected a rule-based algorithm suggested in \cite{wei2019eda}, where the noun and verb tokens are replaced with semantically similar tokens. In the feature extraction module, we used a pre-trained ResNet \cite{he2015deep} network for $net_{img}$ the classification head of the model was removed, and the image embedding size $d_i$ was set to $512$. For the text embedding we used a pre-trained BERT \cite{devlin2019bert} language model provided in \cite{wolf-etal-2020-transformers}. The final size of the sentence embedding $d_t$ was obtained as $768$ by summing the last four hidden states of each token. The internal parameters of the image and text encoders were kept fixed during the training of the hash learning module. In the hash learning module, the networks $f$ (image hashing) and $g$ (text hashing) are fully connected networks with $3$ layers and a batch normalization layer after the second layer. For the discriminator $D$ we selected a simple $2$ layer fully connected network. \textit{ReLU} was used as a default activation function for all layers except for the last layers of image hashing and text hashing networks that use the \textit{tanh} activation function. The hyperparameters $\alpha$, $\beta$, $\gamma$ were set to $\alpha = 0.01$, $\beta = 0.001$, $\gamma = 0.01$ based on a grid search strategy. Both intra-modal weight coefficients $\lambda_1$ and $\lambda_2$ were set to $1$ (see \eqref{eq:contrastive-loss}), while the batch size was set to 256 and the total number of training epochs was selected as $100$. The initial learning rate was set to $0.0001$ and decreased by one fifth every $50$ epochs. The Adam optimizer was chosen for $f$, $g$, and $D$ networks. 
We compared the performance of the proposed method with three state of the art methods: i) supervised multi-task consistency-preserving adversarial hashing (CPAH) \cite{xie2020multi} that separates the feature representations into modality-specific and modality-common features, exploiting label information to learn consistent modality-specific representations; ii) unsupervised method DJSRH \cite{su2019deep}; iii) unsupervised method JDSH \cite{liu2020joint}. For a fair comparison, we trained all models under the same experimental setup. Results of each method are provided in terms of: i) mean average precision (mAP) and ii) precision. The mAP performance was assessed on top-20 retrieved images (denoted as mAP@20), while precision was evaluated by varying the number $K$ of retrieved images in the range of [1-100].



\begin{figure}[t]
  \centering
  \includegraphics[width=\linewidth]{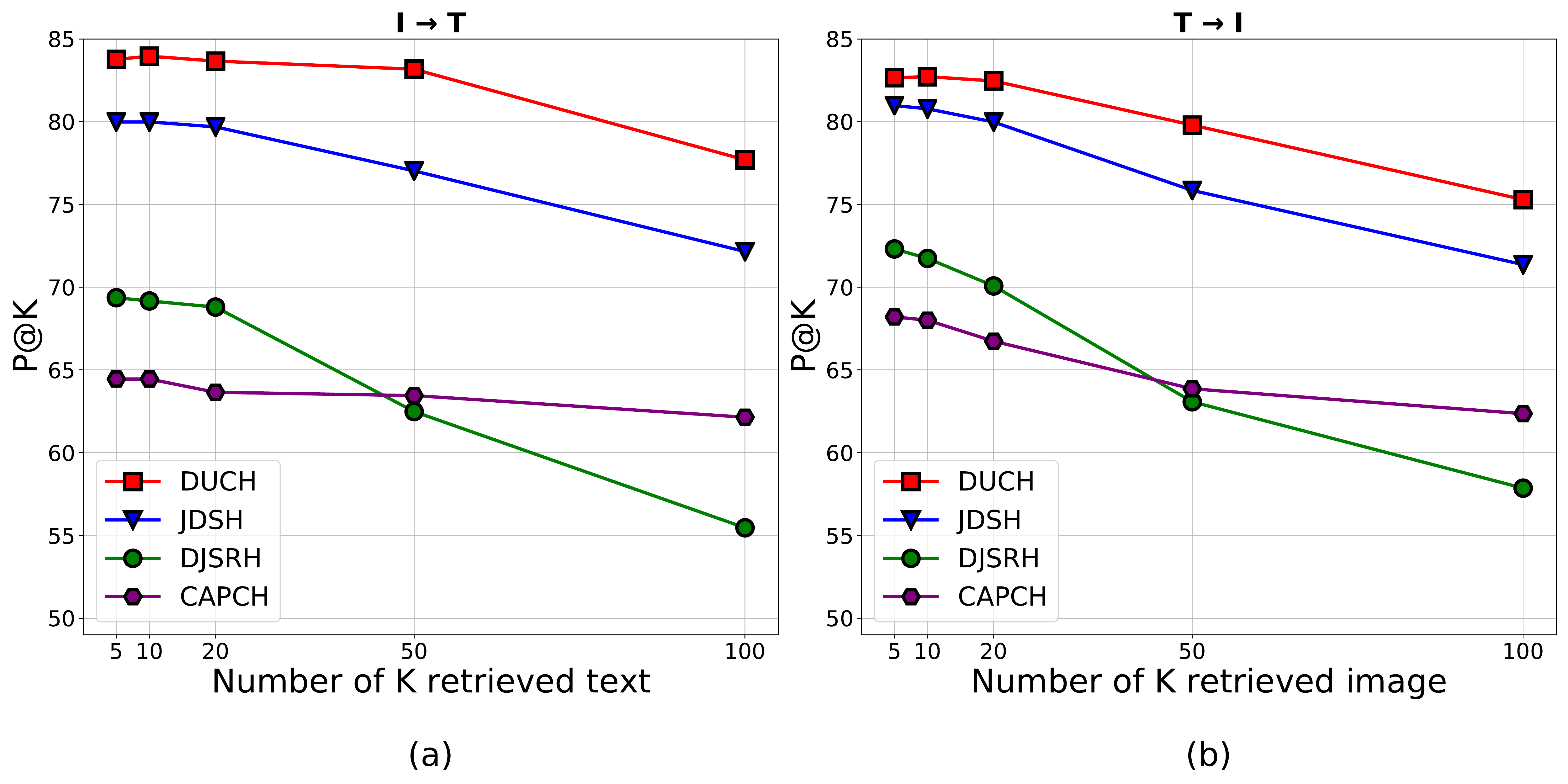}
  \caption{Precision versus number of retrieved images (P@$K$) obtained when $B$=64 for (a) image-to-text ($I \to T$) and (b) text-to-image ($T \to I$) retrieval tasks.}
  \label{fig:precision-at-k}
\end{figure}

In our experiments, we initially analyzed the impact of hash code lengths on retrieval performance. Table \ref{tab:retrieval-performance} shows the retrieval results obtained by the proposed DUCH and the state-of-the-art methods. From the table, one can see that in general, when the number of the hash bits increases for image-to-text and text-to-image retrieval tasks, the mAP@20 obtained by all methods increases. In addition, the proposed DUCH outperforms all the baselines, especially when hash codes with short lengths are considered. As an example, when $B$=16 the mAP@20 is $29.9\%$ higher than that of JDSH (which is the second-best performing method according to our results). The mAP@20 obtained by DJSRH and CPAH are $27.4\%$ and $25.6\%$ smaller than DUCH, respectively. The differences in mAPs are reduced when the higher lengths of the hash codes are considered. We would like to note that DJSRH and JDSH are unsupervised methods that have several hyperparameters and are very sensitive to tweaking. We also analyzed the performance of the proposed DUCH under different values of $K$ (the number of retrieved images) and compared it to the baseline methods. Precision versus the number $K$ of retrieved images when $B$=64 are shown in Fig. \ref{fig:precision-at-k}. From the figure one can observe that the precision obtained by the DUCH (in each value of $K$) is higher than the baseline methods. This shows that the success of the proposed DUCH in efficiently mapping the semantic information into discriminative hash codes.

\begin{table}
\setlength{\tabcolsep}{5pt}
\centering

\caption{The mAP@20 results for image-to-text ($I \to T$) and text-to-image ($T \to I$) retrieval tasks.}
\small
\label{tab:retrieval-performance}
\def\arraystretch{1.0}
\begin{tabular}{c|l|cccc} 
\hline
Task & Method & $B$=16 & $B$=32 & $B$=64 & $B$=128  \\ 
\hline\hline
\multirow{4}{*}{$I\to T$}        

& CPAH \cite{xie2020multi}      & 0.428 & 0.587 &0.636 & 0.696 \\
& DJSRH \cite{su2019deep}       & 0.411 & 0.665 & 0.688 & 0.722  \\
& JDSH \cite{liu2020joint}      & 0.385 & 0.720 & 0.796 & 0.815  \\
& DUCH \textit{(ours)}          & \textbf{0.684} & \textbf{0.791} & \textbf{0.836} & \textbf{0.829}  \\
\hline\hline

\multirow{4}{*}{$T\to I$}

& CPAH \cite{xie2020multi}      & 0.452 & 0.598 & 0.667 & 0.706  \\
& DJSRH \cite{su2019deep}       & 0.422 & 0.685 & 0.705 & 0.733  \\
& JDSH \cite{liu2020joint}      & 0.418 & 0.751 & 0.799 & 0.815 \\
& DUCH \textit{(ours)}          & \textbf{0.697} & \textbf{0.780} & \textbf{0.824} & \textbf{0.826}  \\
\hline
\end{tabular}
\end{table}

To analyze the influence of each objective, we designed different configurations by excluding individual objectives from DUCH. In detail, we compared the original DUCH (that includes all the objectives) with six different configurations of DUCH when we exclude: 1) adversarial objective (denoted as DUCH-NA); 2) quantization objective (denoted as DUCH-NQ); 3) bit balance objective (denoted as DUCH-NB); 4) image and text intra-modal contrastive losses (denoted as DUCH-CL); 5) text intra-modal contrastive loss (denoted as DUCH-CL-I); and 6) image intra-modal contrastive loss (denoted as DUCH-CL-T). The results for different configurations are shown in Table \ref{tab:ablation-study}. The controllable objectives $\mathcal{L}_{adv}$, $\mathcal{L}_Q$ and $\mathcal{L}_{BB}$, from \eqref{eq:loss-overall} only marginally contribute to the overall performance. The highest performance drop (2\%) is observed in the case of DUCH-NQ (which does not include the quantization objective). By analyzing the obtained mAP@20 scores from DUCH-NA and DUCH-NB, one can observe that the Adversarial and Bit balance losses have a less significant impact. This shows that the learned representations with the proposed unsupervised DUCH are consistent, and generated hash codes have independent bits. The intra-modal contrastive losses effectively boost cross-modal unsupervised training compared to the individual single intra-modal contrastive loss (see result of DUCH-CL). Without the intra-modal losses, the performance of the model significantly drops about $8\%$. This shows the importance of the joint use of intra-modal losses to improve the inter-modal contrastive loss. Furthermore, the performance drop in DUCH-CL-I (which does not include text intra-modal contrastive objective) is higher than DUCH-CL-T (which does not contain image intra-modal contrastive objective). This shows that the text intra-modal contrastive objective is more influential on the overall performance than the image intra-modal objective, mainly due to using a modality-specific encoder for the images pre-trained on a different domain.

\begin{table}
\centering
\small
\caption{The mAP@20 results for image-to-text ($I \to T$) and text-to-image ($T \to I$) retrieval tasks under different configurations of the proposed DUCH when $B=64$.}
\label{tab:ablation-study}
\def\arraystretch{1.05}
\begin{tabular}{p{1.75cm}|p{3.45cm}|c|c} 
\hline
Method & Configuration & $I \to T$ & $T \to I$  \\ 
\hline\hline
DUCH         & original & 0.836 & 0.824 \\
DUCH-NA      & excluding $\mathcal{L}_{adv}$: $\alpha = 0$  & 0.831 & 0.822 \\
DUCH-NQ      & excluding $\mathcal{L}_{Q}$: $\beta = 0$     & 0.818 & 0.800 \\
DUCH-NB      & excluding $\mathcal{L}_{BB}$: $\gamma = 0$   & 0.828 & 0.826 \\
DUCH-CL      & excluding $\mathcal{L}_{C_{img}}$, $\mathcal{L}_{C_{txt}}$: $\lambda_1 = 0$, $\lambda_2 = 0$    & 0.758 & 0.765 \\

DUCH-CL-I    & 
excluding $\mathcal{L}_{C_{txt}}$: $\lambda_2 = 0$    & 0.811 & 0.796 \\
DUCH-CL-T    & excluding $\mathcal{L}_{C_{img}}$: $\lambda_1 = 0$   &
0.813 & 0.815 \\
\hline
\end{tabular}
\end{table}

\section{Conclusion}
\label{sec:conclusion}
 
In this paper, a novel deep unsupervised contrastive hashing (DUCH) method has been proposed for cross-modal image-text retrieval in RS. The proposed DUCH exploits a multi-objective loss function to learn a cross-modal representation in an unsupervised fashion. In detail, we have introduced a contrastive objective function that considers both inter- and intra-modal similarities. Furthermore, we have presented an adversarial objective function that assists in generating modality-invariant representations. We have demonstrated the effectiveness of the proposed inter-modality contrastive losses through the experimental results and show the superiority of the proposed DUCH over existing state-of-the-art methods. As a final remark we would like to note that through the experimental results we also observed that the intra-modal contrastive loss is less effective when the deep features are obtained from a modality-specific encoder pre-trained on a different domain. To address this problem, as a future work, we plan for end-to-end training of the feature extraction module and fine-tuning the modality-specific encoders.

\section{Acknowledgments}
\label{sec:acknowledgments}

This work is funded by the European Research Council (ERC) through the ERC-2017-STG BigEarth Project under Grant 759764 and by the German Ministry for Education and Research as BIFOLD - Berlin Institute for the Foundations of Learning and Data (01IS18025A).


\bibliographystyle{IEEEbib}
\bibliography{refs}

\end{document}